\documentclass{article}

\usepackage[preprint]{neurips_2026}

\usepackage[utf8]{inputenc} 
\usepackage[T1]{fontenc}    
\usepackage{hyperref}       
\usepackage{url}            
\usepackage{booktabs}       
\usepackage{amsfonts}       
\usepackage{nicefrac}       
\usepackage{microtype}      
\usepackage{xcolor}         
\usepackage{amsmath}
\usepackage{multirow}
\usepackage{bm}
\usepackage{booktabs}
\usepackage{graphicx} 
\usepackage{wrapfig}
\usepackage{fontawesome5}
\usepackage{graphicx}
\usepackage{booktabs}
\usepackage{multirow}
\usepackage{caption}
\title{Memory Attention}

\author{%
  Jiale Kang \\
  \texttt{kangjiale827@gmail.com} \\
}

\begin{document}
\maketitle
{
\vspace{-11mm}
\begin{center}
        \fontsize{9pt}{\baselineskip}\selectfont
        {\faIcon{github} \tt\href{https://joluck.github.io/memory-attention}{\textbf{memory-attention}}}
        \vspace{2mm}
\end{center}
}
\begin{abstract}
Language models typically construct attention values from contextual hidden states, even when some of their content may be reusable across contexts. We investigate whether token-indexed memory can replace the dedicated value projection when complemented by contextual information. We propose \textbf{Memory Attention (MA)}, which forms values by combining layer-specific token memory with contextual keys. The memory supplies token-specific representations, while the keys preserve context dependence. At inference, normalization can be folded into the memory tables, reducing value construction to lookup and addition. Token-indexed retrieval also enables CPU offloading with prefetching, reducing GPU parameter storage. Under matched training token budgets and with additional memory parameters, experiments across attention configurations show improved language modeling and average downstream performance.
\end{abstract}
\section{Introduction}
\begin{figure}[ht]
    \centering
    \vspace{-1mm}
    \includegraphics[width=\linewidth]{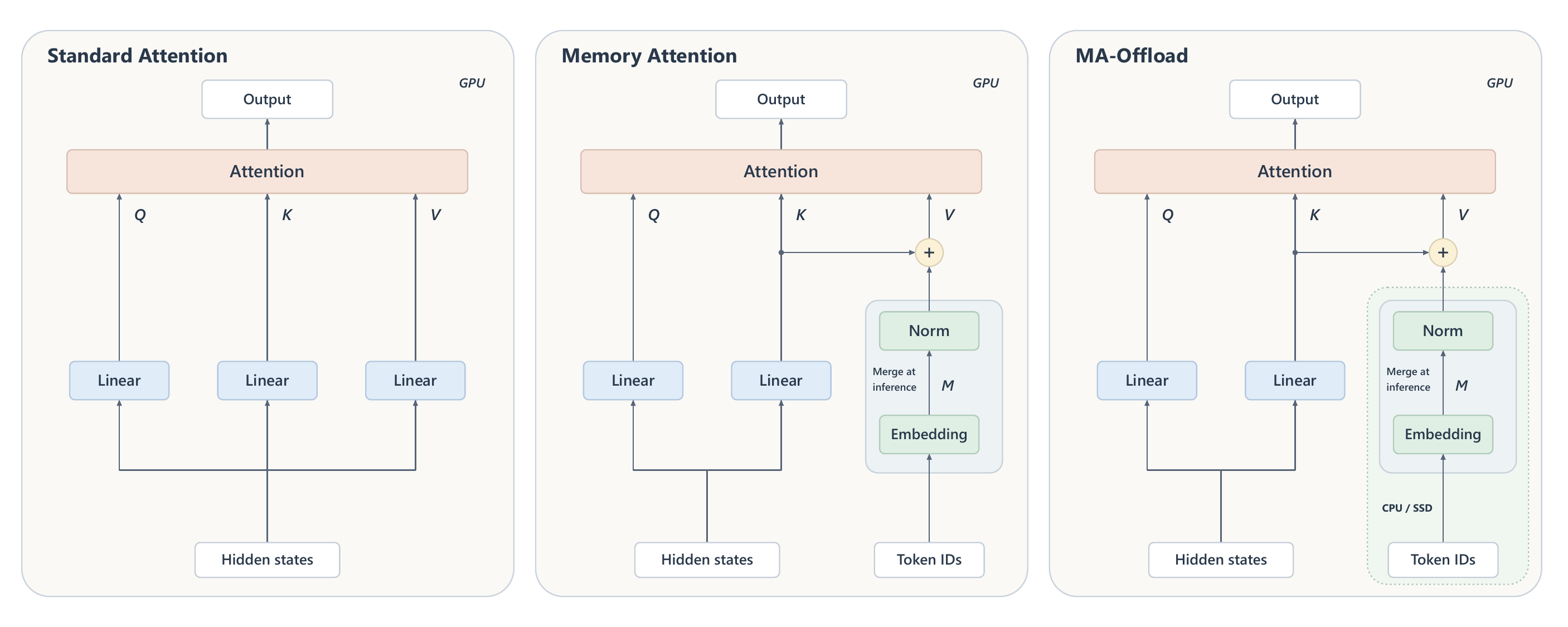}
    \vspace{-6.5mm}
    \caption{
    Standard Attention, Memory Attention, and MA-Offload.
    Memory Attention replaces the value projection with
    ${\bf V}={\bf K}+\operatorname{Norm}({\bf M})$,
    where ${\bf M}$ is retrieved by token ID.
    At inference, normalization is merged into the embedding table,
    which MA-Offload stores on CPU or SSD.
    Additional components used in experiments, including RoPE, are omitted.
    }
    \label{fig:trajectory}
    \vspace{-2mm}
\end{figure}

Improving language model quality under limited computation and GPU memory remains a central challenge. One approach is to expand model capacity through lookup-based representations, which provide additional learned parameters without proportional increases in dense computation. Recent methods, including Value Embedding~\citep{koszarsky2024valueembeddings}, DeepEmbed~\citep{peng_bo_2021_5196578}, Per-Layer Embeddings (PLE)~\citep{gemma2025gemma3n}, STEM~\citep{sadhukhan2026stem}, and Engram~\citep{cheng2026engram}, explore this direction through learned embeddings and memory modules. These developments raise a complementary question: can explicit memory replace part of an existing computation, rather than only supplement it?

Self-attention provides a natural setting for this question. Viewed as a content-addressable memory, attention uses query--key interactions to determine which information to retrieve and values to supply the content to be aggregated. Standard attention constructs values from contextual hidden states through a dedicated linear projection~\citep{vaswani2017attention}. We hypothesize that token-specific representations shared across contexts can supply part of this content, while an existing contextual representation supplies the remainder. In particular, keys already encode contextual information for attention. Can they also provide the contextual component of values, allowing token-indexed memory to replace the dedicated value projection?

We propose \textbf{Memory Attention (MA)}, which forms values by adding learned, layer-specific token memory to contextual keys. Each layer maintains its own embedding table, allowing the same token to have distinct stored representations at different depths. The retrieved embeddings contribute token-specific content, while the keys preserve context dependence. Unlike value embeddings that supplement projected values, MA removes the independent value projection and reuses the key representation. It introduces explicit memory directly into the value pathway while preserving the standard attention weighting and aggregation rules.

This representation changes both value construction and parameter placement. MA replaces a dense value projection with embedding lookup, normalization, and addition. At inference, normalization can be precomputed and folded into the memory tables, leaving only lookup and addition for online value construction. Although the tables increase total parameter storage, their entries are addressed solely by token IDs and layer indices. We exploit this property in \textbf{MA-Offload}, which stores the tables in CPU memory and prefetches the required entries before the corresponding layer uses them. This allows additional memory capacity to reside outside the GPU, with a latency cost determined by retrieval, transfer, and their overlap with computation. The additive representation also permits values to be reconstructed from retained keys and token memory. We analyze this possibility as \textbf{MA-Recall}, an extension that trades repeated retrieval and reconstruction for reduced persistent value-cache storage.

We evaluate MA through language modeling, downstream tasks, retrieval, and inference measurements. Across the evaluated attention configurations, MA improves both language modeling perplexity metrics and average downstream accuracy under matched training token budgets, while introducing additional memory parameters. Retrieval experiments also show gains within the training context window and at twice its length. Inference measurements demonstrate that CPU offloading can accommodate the larger MA parameterization with lower GPU parameter storage than the standard baseline, while introducing decoding overhead. These results characterize the quality and efficiency of the combined design; they do not isolate the contribution of its structure from the increase in parameter capacity.

Our main contributions are as follows:
\begin{itemize}
    \item We propose \textbf{Memory Attention}, which combines layer-specific token memory with contextual keys to replace the dedicated value projection while preserving the standard attention weighting and aggregation rules.
    \item We develop and evaluate \textbf{MA-Offload}, which exploits token-indexed retrieval to place memory tables in CPU memory and prefetch the required entries. We further analyze \textbf{MA-Recall} as a potential extension for reconstructing values without a persistent value cache.
    \item We evaluate model quality and inference behavior, and analyze the trade-offs among parameter capacity, value-construction arithmetic, GPU parameter storage, memory traffic, and persistent cache storage.
\end{itemize}

\section{Background}
\label{sec:background}

We review self-attention and three mechanisms for
incorporating token-indexed memory.
Layer indices are omitted throughout; embedding tables
and transformations may differ across layers.

\paragraph{Self-Attention.}
Given hidden states $\mathbf{X}$, self-attention computes
queries, keys, and values through linear projections:
\begin{equation}
    \mathbf{Q}=\mathbf{X}\mathbf{W}_Q,\quad
    \mathbf{K}=\mathbf{X}\mathbf{W}_K,\quad
    \mathbf{V}=\mathbf{X}\mathbf{W}_V.
\end{equation}
For a single head, the attention output is
\begin{equation}
    \operatorname{Attn}(\mathbf{X})
    = \operatorname{softmax}\!\left(
        \frac{\mathbf{Q}\mathbf{K}^{\top}}{\sqrt{d_k}}
        + \mathbf{C}
    \right)\mathbf{V},
\end{equation}
where $d_k$ is the key dimension and $\mathbf{C}$ is
the causal mask.
Query--key interactions determine the aggregation
weights, while the values supply the content to be
aggregated.
All three representations are derived from the
current hidden states.

\paragraph{Value Embeddings.}
We use $\mathbf{E}$ to denote a learnable embedding
table and $\mathbf{E}[i]$ to denote an embedding lookup
that retrieves its $i$-th row.
Value embeddings supplement projected values with
token-specific representations from
$\mathbf{E}\in\mathbb{R}^{N\times d_v}$:
\begin{equation}
    \mathbf{v}_{e,t}=\mathbf{E}[s_t],
    \qquad
    \widetilde{\mathbf{v}}_t
    = (1-\lambda)\mathbf{v}_t
      + \lambda\mathbf{v}_{e,t}.
\end{equation}
where $s_t$ is the token ID and $\lambda$ controls
the embedding contribution.
This combines contextual values with token-specific
representations while retaining the original value
projection.

\paragraph{Per-Layer Embeddings.}
Per-Layer Embeddings (PLE) introduce a learnable
embedding table $\mathbf{E}\in\mathbb{R}^{N\times d_e}$
at each layer, where $d_e$ is the embedding dimension.
Omitting layer indices, normalization, and scaling,
the update is
\begin{align}
    \mathbf{e}_t
        &= \mathbf{E}[s_t]
           + \mathbf{P}\mathbf{x}_t^{(0)}, \\
    \Delta\mathbf{h}_t
        &= \mathbf{W}_o
           \left[
               \operatorname{GELU}\!\left(
                   \mathbf{W}_g\mathbf{h}_t
               \right)
               \odot \mathbf{e}_t
           \right],
\end{align}
where $\mathbf{x}_t^{(0)}$ is the input embedding
and $\mathbf{h}_t$ is the hidden state at the
injection point.
The lookup and projected input embedding provide
token-specific information, while the hidden-state
gate modulates its contribution to the update.

\paragraph{Engram.}
Engram retrieves learned memory using local
$n$-gram patterns.
Let $\mathbf{E}\in\mathbb{R}^{H\times d_m}$ denote
a learnable memory table with $H$ entries of
dimension $d_m$.
Abstracting the multiple lookups in the full
architecture, a single retrieval is
\begin{equation}
    \mathbf{m}_t
    = \mathbf{E}\!\left[
        h(s_{t-n+1},\ldots,s_t)
    \right],
\end{equation}
where $h$ maps a local $n$-gram to a table index.
The retrieved memory is integrated through contextual
gating.
Because lookup addresses depend only on token IDs,
retrieval can be prefetched before the corresponding
hidden states become available.
This allows memory capacity to grow without a
proportional increase in dense computation.

\paragraph{DeepEmbedAttention (DEA)~\cite{peng_bo_2021_5196578} }incorporates token-specific embeddings into attention by multiplicatively modulating keys and values derived from contextual hidden states. It combines low-dimensional key and value projections with up-projections and token-indexed embedding tables, enabling a compact KV cache. Introduced as an attention component for hybrid RWKV-7s models, DEA illustrates how lexical information stored in embedding tables can complement contextual representations. The author also suggested replacing key and value computation with direct token-indexed table lookups, substantially reducing KV cache requirements.

\paragraph{Bank of Values.}
Recent work~\cite{he2026value} investigates whether value
vectors in deeper Transformer layers require contextual
information from the residual stream. The authors find
that context-free, token-specific values can improve
model performance, while adding contextual components
provides limited additional benefit in deeper layers.
Based on this observation, they propose Bank of Values
(BoV), which replaces contextual value projections
in the last third of layers with learnable token-indexed
value tables. These findings highlight the potential
of token-specific representations as an alternative
source of attention values.
Unlike BoV, which directly replaces values with
context-free embeddings in selected layers, MA
combines token memory with contextual keys to
construct values without an independent value projection.

\section{Memory Attention}
\label{sec:method}

Memory Attention (MA) replaces the dedicated value projection with a combination of contextual keys and layer-specific token memory. We first define this construction and its integration into attention, then analyze its computational and storage costs. We omit layer indices unless needed; each MA layer has its own memory table and learned transformations.

\subsection{Memory-Based Value Construction}

Let $\mathbf{X}\in\mathbb{R}^{T\times d}$ denote the hidden states of a sequence with token IDs $\mathbf{s}=(s_1,\ldots,s_T)$. Let $H_{\mathrm{KV}}$ denote the number of key/value heads and $d_h$ their head dimension. We assume matching key and value dimensions and write $d_v=H_{\mathrm{KV}}d_h$ for the total dimension across key/value heads.

Each MA layer maintains a learnable memory table $\mathbf{E}\in\mathbb{R}^{N\times d_v}$, where $N$ is the vocabulary size. The token memory is
\begin{equation}
    \mathbf{M}
    = \operatorname{Norm}(\mathbf{E}[\mathbf{s}])
    \in\mathbb{R}^{T\times d_v},
    \label{eq:memory_lookup}
\end{equation}
where $\mathbf{E}[\mathbf{s}]$ retrieves the rows indexed by the input tokens. $\operatorname{Norm}$ applies RMSNorm independently to each token vector within each key/value head. The memory tables and normalization parameters are learned jointly with the rest of the model. Within a layer, the same token retrieves the same memory vector across positions and contexts; separate tables allow different layers to learn distinct representations.

MA computes queries and keys from the hidden states, then constructs values by addition:
\begin{equation}
    \mathbf{Q}=\mathbf{X}\mathbf{W}_Q,\qquad
    \mathbf{K}=\mathbf{X}\mathbf{W}_K,\qquad
    \mathbf{V}=\mathbf{K}+\mathbf{M}.
    \label{eq:memory_values}
\end{equation}
The memory supplies token-specific representations, while the keys provide dynamically computed information from the current hidden states. These hidden states incorporate contextual information and intermediate representations accumulated through preceding layers. The key contribution therefore reflects both the current context and the computation performed up to the current depth. Layer-specific memory complements this contribution with stored token representations. Unlike supplementing projected values with token embeddings, MA removes the independent value projection $\mathbf{W}_V$ and jointly learns the key projection and memory table to support value construction.

For attention with rotary positional embeddings (RoPE), positional transformations are applied to the queries and keys used for attention scoring:
\begin{equation}
    \mathbf{Q}^{R}=\operatorname{RoPE}(\mathbf{Q}),\qquad
    \mathbf{K}^{R}=\operatorname{RoPE}(\mathbf{K}),\qquad
    \mathbf{V}=\mathbf{K}+\mathbf{M}.
    \label{eq:ma_rope}
\end{equation}
Values are constructed from $\mathbf{K}$ before this transformation, while $\mathbf{Q}^{R}$ and $\mathbf{K}^{R}$ are used to compute attention weights. The resulting values are passed to the original attention operator, leaving its weighting and aggregation rules unchanged.

\subsection{Contextual and Token-Specific Value Decomposition}
\label{sec:ma_decomposition}

Memory Attention can be interpreted as key--value weight sharing
augmented with a token-dependent embedding bias.
While standard attention constructs keys and values through
independent projections, MA reuses the key projection to
construct values:
\begin{equation}
    \mathbf{V}
    = \underbrace{\mathbf{X}\mathbf{W}_K}
      _{\text{Shared contextual projection}}
    + \underbrace{\operatorname{Norm}(\mathbf{E}[\mathbf{s}])}
      _{\text{Token-dependent bias}}.
\end{equation}
Unlike a conventional linear bias shared across all tokens,
the memory bias depends on token identity and varies across
layers. This formulation removes the independent value
projection while introducing token-specific representations
that complement the shared contextual features.

The resulting attention output admits a decomposition into
contextual and memory contributions:
\begin{equation}
\begin{aligned}
    \mathbf{O}_{\mathrm{MA}}
    &= \mathbf{A}(\mathbf{K}+\mathbf{M}) \\
    &= \underbrace{\mathbf{A}\mathbf{K}}
       _{\text{Contextual retrieval}}
     + \underbrace{\mathbf{A}\mathbf{M}}
       _{\text{Memory retrieval}},
\end{aligned}
\label{eq:ma_decomposition}
\end{equation}
where $\mathbf{A}$ denotes the attention weights and
$\mathbf{K}$ represents keys before positional transformations.
The first term aggregates contextual representations derived
from hidden states, while the second retrieves token-specific
information under the same attention distribution.

Although each memory embedding is fixed for a given token
within a layer, its aggregated contribution remains
context-dependent. For a query at position $t$, the memory
contribution can be expressed as
\begin{equation}
    \mathbf{o}^{\mathrm{mem}}_t
    = \sum_{j \leq t} a_{tj}\mathbf{m}_{s_j},
\end{equation}
where $a_{tj}$ depends on the query and contextual keys.
Thus, the same token memory can contribute differently
across contexts through attention-weighted aggregation,
without requiring an independent contextual value projection.

This decomposition highlights the complementary roles of
the two components: keys provide dynamically computed
contextual information, while token memory supplies an
additional source of learned content. Both are jointly
optimized to support value aggregation. Rather than
constructing values through a separate dense projection,
MA combines a shared contextual representation with
token-indexed memory, offering an alternative allocation
of model capacity between computation and storage.

\subsection{Computation and Parameter Storage}
\label{sec:cost}

We compare the value-construction costs of MA and standard attention across $L$ replaced layers. Let $S$ denote the number of newly processed tokens: $S=BT$ for prefill with batch size $B$ and sequence length $T$, and $S=B$ for one decoding step. We assume bias-free projections and denote the number of learned normalization parameters per layer by $p_{\mathrm{norm}}$. Shared components, including the key projection, are excluded from this comparison.

\paragraph{Parameter capacity.}
Standard attention uses $Ldd_v$ parameters for the value projections. MA replaces them with $L(Nd_v+p_{\mathrm{norm}})$ memory and normalization parameters. When $N>d$, this increases parameter capacity and total parameter storage. The additional capacity is accessed through token lookup rather than a dense projection.

\paragraph{Training computation.}
The standard value projection requires approximately $6LSdd_v$ FLOPs for the forward pass and the input and weight gradients. MA instead requires $\mathcal{O}(LSd_v)$ arithmetic for head-wise normalization, addition, and gradient accumulation over accessed memory entries. This comparison excludes optimizer updates, gradient-buffer initialization, and memory traffic. In particular, lookup-based computation does not by itself imply sparse gradient storage or sparse optimizer states.

\paragraph{Inference computation.}
At inference, both the memory tables and normalization parameters are fixed. Because normalization acts independently on each retrieved row, it can be folded into the tables:
\begin{equation}
    \overline{\mathbf{E}}[i]
    = \operatorname{Norm}(\mathbf{E}[i]),
    \qquad i=1,\ldots,N.
    \label{eq:folded_memory}
\end{equation}
Online value construction then becomes
\begin{equation}
    \mathbf{V}
    = \mathbf{K}+\overline{\mathbf{E}}[\mathbf{s}],
\end{equation}
requiring only lookup and element-wise addition. Table~\ref{tab:value_cost} summarizes the parameter and inference arithmetic costs.

\begin{table}[t]
    \centering
    \small
    \setlength{\tabcolsep}{4pt}
    \begin{tabular}{lcc}
        \toprule
        Metric & Standard & MA \\
        \midrule
        Value construction
            & $\mathbf{X}\mathbf{W}_V$
            & $\mathbf{K}+\overline{\mathbf{E}}[\mathbf{s}]$ \\
        Trainable parameters
            & $Ldd_v$
            & $L(Nd_v+p_{\mathrm{norm}})$ \\
        Stored inference parameters
            & $Ldd_v$
            & $LNd_v$ \\
        Online value-construction FLOPs
            & $\approx 2LSdd_v$
            & $LSd_v$ \\
        \bottomrule
    \end{tabular}
    \caption{
        Value-construction costs across $L$ replaced layers. Shared components, including the key projection, are excluded. Each multiplication or addition counts as one FLOP. Lookup overhead and memory traffic are excluded. Inference costs apply to newly processed tokens and do not include reconstruction of historical values.
    }
    \label{tab:value_cost}
\end{table}

The reduction in online value-construction arithmetic is approximately
\begin{equation}
    \Delta F_V \approx LSd_v(2d-1).
\end{equation}
If $F_{\mathrm{common}}$ denotes the FLOPs of all unchanged components for the same workload, the complete-model FLOP reduction is
\begin{equation}
    \eta_{\mathrm{model}}
    \approx
    \frac{LSd_v(2d-1)}
         {F_{\mathrm{common}}+2LSdd_v}.
\end{equation}
This arithmetic reduction does not directly determine latency, which also depends on memory access, kernel execution, and synchronization.

\subsection{MA-Offload}
\label{sec:ma_offload}

The folded memory tables need not remain on the GPU. Their lookup addresses depend only on token IDs and layer indices, so the required entries can be identified before the corresponding hidden states are available. \textbf{MA-Offload} stores the folded tables in CPU memory and prefetches the required vectors to the GPU, allowing retrieval and transfer to overlap with model computation.

Let $b_w$ denote the number of bytes per stored memory element. Offloading moves
\begin{equation}
    P_{\mathrm{offloaded}}=b_wLNd_v
\end{equation}
bytes of memory-table parameters from GPU to CPU storage. It does not change the total parameter count or the value-construction arithmetic. The GPU still requires temporary lookup and transfer buffers.

Without caching or deduplication, the logical host-to-device payload for $S$ newly processed tokens is
\begin{equation}
    D_{\mathrm{Offload}}=b_wLSd_v.
    \label{eq:offload_traffic}
\end{equation}
With a conventional KV cache, historical values remain cached, so each decoding step retrieves memory only for the new tokens. The resulting payload is $b_wLBd_v$ bytes per step.

MA-Offload reduces GPU parameter residency while retaining the historical key and value cache. Its latency cost depends on CPU lookup performance, transfer bandwidth, synchronization, and the overlap available in the execution schedule. Consequently, moving the tables off the GPU reduces parameter storage but does not guarantee a reduction in either total GPU memory usage or inference latency.

\subsection{MA-Recall: An Extension for Value Reconstruction}
\label{sec:ma_recall}
The additive value representation also permits reconstruction of historical values. We analyze this possibility as \textbf{MA-Recall}, an extension that avoids a persistent value cache by retaining sufficient key information and the historical token IDs. MA-Recall is not included in the inference measurements reported in this work.
Given historical content keys and token IDs, values are recovered as
\begin{equation}
\mathbf{V}{1:T}
= \mathbf{K}{1:T}
+ \overline{\mathbf{E}}[\mathbf{s}_{1:T}].
\label{eq:ma_recall}
\end{equation}
A practical implementation must also support the keys required for attention scoring. With standard RoPE, a cached rotated key can in principle be mapped back to its content representation using the inverse positional rotation. Alternatively, content keys can be retained and rotated when needed. Either choice introduces positional-transformation work that must be included in implementation-level cost measurements. Retaining both representations would require additional cache storage.
Let $b_c$ denote the number of bytes per cached key or value element. For equal key and value dimensions and precision, a conventional content KV cache occupies $2b_cLBTd_v$ bytes. Retaining one key representation per historical position reduces this component to $b_cLBTd_v$ bytes, a reduction of 50\%. This accounting excludes token IDs, position metadata, temporary reconstruction buffers, and any additional positional-key storage. Blockwise reconstruction could avoid materializing the complete value history on the GPU.
The storage reduction comes at the cost of repeated memory access and reconstruction. Let $R$ denote the number of historical positions whose values are reconstructed per sequence in a decoding step, assuming the same count across layers. Reconstruction requires approximately $LBRd_v$ additions, excluding any positional transformations. For full causal attention, $R=T$; for restricted attention patterns, $R$ is the number of positions actually reconstructed.
When memory tables are also CPU-offloaded, the corresponding logical transfer payload is \begin{equation} D_{\mathrm{Recall}}=b_wLBRd_v, \end{equation} without caching or deduplication. MA-Recall therefore trades persistent cache storage for additional retrieval, transfer, and reconstruction work. Its practical benefit depends on the attention pattern, cache representation, memory placement, and reconstruction implementation.

\section{Experiments}
\label{sec:experiments}

We evaluate MA in terms of language modeling and downstream performance, training efficiency, retrieval, and inference behavior. The quality evaluations examine whether token memory and shared keys can effectively replace the dedicated value projection. The efficiency evaluations characterize the computational and parameter-storage trade-offs of GPU-resident and CPU-offloaded memory tables.

\subsection{Experimental Setup}

\paragraph{Training configurations.}
We train models on NVIDIA H800 GPUs using the \texttt{flash-linear-attention}~\citep{yang2024fla} framework. The backbone consists of attention blocks and gated MLPs, with rotary positional embeddings (RoPE) and RMSNorm. Detailed pretraining settings are provided in Appendix~\ref{sec:pretraining_settings}, and model configurations are summarized in Table~\ref{tab:main_results}. Within each configuration, Standard and MA use matched training token budgets, while MA introduces additional memory parameters.


\paragraph{Evaluation benchmarks.}
We evaluate language modeling and downstream performance in a zero-shot setting using \texttt{lm-evaluation-harness}~\citep{eval-harness}. We report perplexity on LAMBADA~\citep{paperno2016lambada} and WikiText~\citep{merity2016pointer}, together with LAMBADA word-prediction accuracy. Downstream benchmarks include ARC-Easy and ARC-Challenge~\citep{clark2018think}, HellaSwag~\citep{zellers2019hellaswag}, PIQA~\citep{bisk2020piqa}, WinoGrande~\citep{sakaguchi2021winogrande}, and OpenBookQA~\citep{mihaylov2018can}. We use normalized accuracy for ARC-Challenge and HellaSwag and accuracy for the remaining tasks. The average score is the unweighted mean of the seven accuracy-based scores, excluding perplexity.


\subsection{Language Modeling}
\label{sec:main_results}

\begin{table*}[t]
\centering
\scriptsize
\renewcommand{\arraystretch}{1.2}
\setlength{\tabcolsep}{4pt}
\resizebox{\textwidth}{!}{%
\begin{tabular}{llrcc|cccccccc}
\toprule
\multirow{2}{*}{\textbf{Backbone}}
& \multirow{2}{*}{\textbf{Variant}}
& \multirow{2}{*}{\textbf{Params.}}
& \textbf{Lamb.}
& \textbf{WikiText}
& \textbf{Lamb.}
& \textbf{ARC-E}
& \textbf{ARC-C}
& \textbf{HellaSwag}
& \textbf{PIQA}
& \textbf{WinoGrande}
& \textbf{OBQA}
& \textbf{Avg.} \\
& & &
$\mathrm{PPL}\downarrow$
& $\mathrm{PPL}\downarrow$
& $\mathrm{Acc.}\uparrow$
& $\mathrm{Acc.}\uparrow$
& $\mathrm{Acc.}_{n}\uparrow$
& $\mathrm{Acc.}_{n}\uparrow$
& $\mathrm{Acc.}\uparrow$
& $\mathrm{Acc.}\uparrow$
& $\mathrm{Acc.}\uparrow$
& $\uparrow$ \\
\midrule
\multicolumn{13}{l}{
\textit{24 layers, hidden size 1,024; 10B training tokens; 0.5M tokens per batch}
} \\
\midrule
\multirow{2}{*}{MHA}
& Standard & 373M
& 44.35 & 31.55 & 31.59 & 53.91 & \textbf{26.96}
& 36.86 & 65.18 & 49.09 & \textbf{21.20} & 40.68 \\
& Memory & 1,135M
& \textbf{42.78} & \textbf{28.64}
& \textbf{32.25} & \textbf{56.44} & 25.85
& \textbf{38.41} & \textbf{66.00} & \textbf{50.36}
& 20.40 & \textbf{41.39} \\
\addlinespace
\multirow{2}{*}{GQA}
& Standard & 349M
& 50.46 & 31.71 & 30.86 & 54.21 & 27.47
& 36.59 & \textbf{65.83} & 48.93 & \textbf{22.40} & 40.90 \\
& Memory & 729M
& \textbf{43.68} & \textbf{29.45}
& \textbf{30.97} & \textbf{56.27} & \textbf{27.65}
& \textbf{38.07} & 65.18 & \textbf{51.62}
& 20.80 & \textbf{41.51} \\

\addlinespace
\multirow{2}{*}{MQA}
& Standard & 336M
& 50.38 & 31.84 & 31.32 & 54.17 & \textbf{26.88} & 35.98 & 64.53 & \textbf{53.28} & \textbf{21.60} & 41.11 \\
& Memory & 526M
&\textbf{45.93} & \textbf{29.82} & \textbf{31.55} & \textbf{55.05} & 26.62 & \textbf{37.55} & \textbf{65.56} & 52.25 & 20.80 & \textbf{41.34} \\
\addlinespace
\multirow{2}{*}{Gate}
& Standard & 398M
& 41.88 & 29.67 & 32.10 & 56.06 & 26.71 & 38.21 & 65.23 & 50.91 & 20.60 & 41.40 \\
& Memory & 1160M
& \textbf{36.60} & \textbf{27.48} & \textbf{33.73} & \textbf{56.69} & \textbf{27.99} & \textbf{39.73} & \textbf{66.65} & \textbf{51.78} & \textbf{23.60} & \textbf{42.88} \\
\midrule
\multicolumn{13}{l}{
\textit{24 layers, hidden size 2,048; 20B training tokens; 1M tokens per batch}
} \\
\midrule
\multirow{2}{*}{MHA}
& Standard & 1,364M
& 18.59 & 21.52 & 40.50 & \textbf{66.67} & 31.31
& 46.98 & \textbf{69.91} & 53.67 & 34.40 & 49.06 \\
& Memory & 2,836M
& \textbf{18.15} & \textbf{20.79}
& \textbf{41.59} & 66.04 & \textbf{33.96}
& \textbf{48.38} & 69.42 & \textbf{54.78}
& \textbf{37.40} & \textbf{50.22} \\
\bottomrule
\end{tabular}%
}
\caption{
Zero-shot language modeling and downstream results at matched training
token budgets. Parameter counts include memory tables and differ within
pairs. Accuracies are percentages; $\mathrm{Acc}_{n}$ denotes normalized
accuracy; Avg.\ averages the seven accuracy scores.
Bold marks the better result in each Standard--Memory pair.
Experiments on additional backbones, including swa and linear
attention, as well as larger models and mixture-of-experts (MoE)
architectures, are ongoing.
}
\vspace{-20pt}
\label{tab:main_results}
\end{table*}

Table~\ref{tab:main_results} shows that MA improves both perplexity metrics and the average accuracy-based score across all three configurations. For example, WikiText perplexity decreases from 31.55 to 28.64 for the smaller MHA model, while LAMBADA perplexity decreases from 50.46 to 43.68 for GQA. The larger model also improves on both language modeling metrics.

The average accuracy gains are 0.71 percentage points for the smaller MHA model, 0.61 for GQA, and 1.16 for the larger MHA model. Improvements vary across tasks: at the larger scale, MA gains 3.00 points on OpenBookQA and 2.65 points on ARC-Challenge, while slightly reducing ARC-Easy and PIQA accuracy. Thus, the aggregate improvements do not imply gains on every benchmark.

These results support the effectiveness of the combined MA design under the reported training budgets. Because MA includes additional memory parameters, they do not isolate the contribution of its structure from the increase in parameter capacity.

These ratios describe the number of training tokens needed to reach the selected loss levels. They do not directly measure wall-clock training speed. MA also reduces value-construction arithmetic, as analyzed in Section~\ref{sec:cost}, but realized training time additionally depends on memory access, gradient handling, and optimizer updates for the larger parameter set.

\paragraph{Retrieval and Context-Length Extrapolation}

We evaluate retrieval within the training context window and under context-length extrapolation using three single-needle NIAH tasks. The evaluated models have 24 layers and a hidden dimension of 1,024 and are trained on 10B tokens with a context length of 2,048. Evaluation lengths are 1,024, 2,048, and 4,096 tokens. The first two fall within the training window, while the last tests extrapolation to twice the training length.

\begin{wraptable}{r}{0.60\textwidth}
\centering
\small
\renewcommand{\arraystretch}{1.2}
\setlength{\tabcolsep}{4pt}
\resizebox{\linewidth}{!}{%
\begin{tabular}{l|ccc|ccc|ccc}
\toprule
\multirow{2}{*}{Variant}
& \multicolumn{3}{c|}{\texttt{niah\_single\_1}}
& \multicolumn{3}{c|}{\texttt{niah\_single\_2}}
& \multicolumn{3}{c}{\texttt{niah\_single\_3}} \\
\cmidrule(lr){2-4}
\cmidrule(lr){5-7}
\cmidrule(lr){8-10}
& 1K & 2K & 4K$^\dagger$
& 1K & 2K & 4K$^\dagger$
& 1K & 2K & 4K$^\dagger$ \\
\midrule
Standard
& 0.998 & 0.844 & 0.414
& 0.856 & 0.914 & 0.232
& 0.624 & 0.704 & 0.130 \\
MA
& \textbf{1.000} & \textbf{0.994} & \textbf{0.510}
& \textbf{0.940} & \textbf{1.000} & \textbf{0.454}
& \textbf{0.982} & \textbf{0.910} & \textbf{0.294} \\
\bottomrule
\end{tabular}%
}
\caption{
Zero-shot single-needle NIAH scores for models
trained with a 2K context window.
1K, 2K, and 4K denote 1,024, 2,048, and 4,096
tokens, respectively.
$^\dagger$ marks extrapolation beyond the training
context length.
Higher scores are better; bold indicates the
better result in each column.
}
\label{tab:long_context}
\end{wraptable}

As shown in Table~\ref{tab:long_context}, MA outperforms Standard across all nine task--length combinations. Within the training window, its mean scores are 97.4\% at 1K and 96.8\% at 2K, compared with 82.6\% and 82.1\% for Standard. The largest individual gain occurs on \texttt{niah\_single\_3} at 1K, where performance increases from 62.4\% to 98.2\%.

At 4K, MA retains an advantage on all three tasks, with a mean score of 41.9\% compared with 25.9\% for Standard. Both models nevertheless degrade substantially beyond the training window. These results demonstrate improved retrieval on the evaluated tasks, including under $2\times$ length extrapolation, but do not establish robust retrieval at longer context lengths.
\subsection{Ablation Studies}
\label{sec:ablation}

We conduct a series of ablation studies to investigate the key
design choices of Memory Attention (MA), including memory
integration, memory placement, normalization, and its
compatibility with attention residuals (AttnRes).
Unless otherwise specified, all models are trained with
24 layers, a hidden size of 1,024, and 10B tokens, using
a batch size of approximately 0.5M tokens.
We report perplexity on LAMBADA and WikiText, along with
accuracy on seven downstream benchmarks.

\paragraph{Memory Integration.}
We first investigate how memory should be incorporated into
the attention mechanism.
As shown in Table~\ref{tab:integration},
replacing the standard value projection with memory alone
($V=M$) degrades the average downstream accuracy from
41.11 to 40.34, suggesting that memory cannot fully replace
input-dependent representations.
Interpolating between the projected values and memory
($V=(1-\lambda)xW+\lambda E[i]$) also provides limited
improvements.
In contrast, additive integration yields substantially
lower perplexity, with $V=xW+M$ achieving a LAMBADA
perplexity of 45.68.
Notably, reusing the keys as values ($V=K+M$) achieves
the best average downstream accuracy of 41.38 and
the lowest WikiText perplexity of 29.82.
These results suggest that memory is most effective as
a complementary component to input-dependent attention
representations, while key--value reuse provides a
simple and effective integration strategy.
\begin{table*}[h]
\centering

\begin{minipage}[t]{0.26\textwidth}
\vspace{0pt}
\centering
\includegraphics[width=\linewidth]{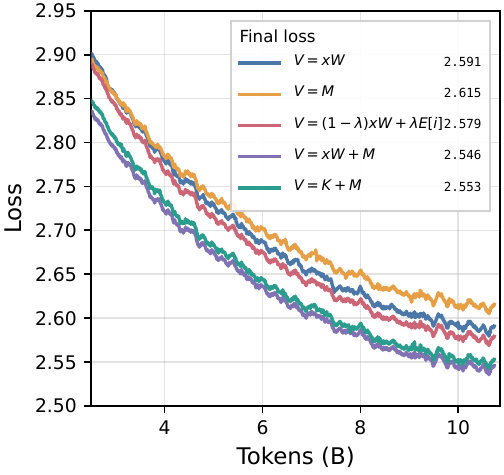}
\end{minipage}%
\hfill
\begin{minipage}[t]{0.7\textwidth}
\vspace{0pt}
\centering
\scriptsize
\renewcommand{\arraystretch}{1.2}
\setlength{\tabcolsep}{4pt}

\resizebox{\linewidth}{!}{%
\begin{tabular}{lrcc|cccccccc}
\toprule
\multirow{2}{*}{\textbf{Configuration}}
& \multirow{2}{*}{\textbf{Params.}}
& \textbf{Lamb.}
& \textbf{WikiText}
& \textbf{Lamb.}
& \textbf{ARC-E}
& \textbf{ARC-C}
& \textbf{HellaSwag}
& \textbf{PIQA}
& \textbf{WinoGrande}
& \textbf{OBQA}
& \textbf{Avg.} \\
& &
$\mathrm{PPL}\downarrow$
& $\mathrm{PPL}\downarrow$
& $\mathrm{Acc.}\uparrow$
& $\mathrm{Acc.}\uparrow$
& $\mathrm{Acc.}_{n}\uparrow$
& $\mathrm{Acc.}_{n}\uparrow$
& $\mathrm{Acc.}\uparrow$
& $\mathrm{Acc.}\uparrow$
& $\mathrm{Acc.}\uparrow$
& $\uparrow$ \\
\midrule
\multicolumn{12}{l}{%
\textit{24 layers, hidden size 1,024; 10B training tokens;
0.5M tokens per batch}
} \\
\midrule
$V=xW$
& 336M & 50.38 & 31.84 & 31.32 & 54.17 & 26.88
& 35.98 & 64.53 & \textbf{53.28} & 21.60 & 41.11 \\
$V=M$
& 526M & 58.36 & 32.94 & 28.84 & 53.45 & \textbf{27.22}
& 35.79 & 65.07 & 52.64 & 19.40 & 40.34 \\
$V=(1-\lambda)xW+\lambda E[i]$& 532M
& 47.87 & 31.12 & 31.38 & 55.47 & 26.79 & 36.72 & 64.80 & 50.75 & 21.00 & 40.99 \\
$V=xW+M$
& 532M & \textbf{45.68} & 29.89 & 31.01
& \textbf{56.02} & 26.79
& \textbf{37.76} & \textbf{65.67} & 51.30 & 20.40 & 41.28 \\
$V=K+M$
& 526M & 45.93 & \textbf{29.82} & \textbf{31.57}
& 55.05 & 26.62
& 37.52 & 65.61 & 52.49 & 20.80 & \textbf{41.38} \\
\bottomrule
\end{tabular}%
}

\captionsetup{
    font=scriptsize,
    labelfont=normalfont,
    justification=raggedright,
    singlelinecheck=false,
    skip=4pt
}

\caption{%
Ablation of memory integration and key--value reuse.
$V=K+M$ and $K=V+M$ denote reusing keys as values and
values as keys, respectively, with an added memory component $M$.
Bold indicates the best result in each metric column.
The left panel shows the corresponding training loss curves.
 }
\label{tab:integration}
\end{minipage}
\vspace{-20pt}
\end{table*}
\paragraph{Memory Placement.}
We next examine where memory should be introduced within
the attention block.
Specifically, we inject memory into the hidden states,
queries, keys, values, or attention outputs while
maintaining shared key--value representations.
Table~\ref{tab:placement} shows that adding memory
to the value representations achieves the best overall
performance, with an average accuracy of 41.38 and
a WikiText perplexity of 29.82.
Injecting memory into hidden states provides a competitive
alternative, whereas modifying queries, keys, or attention
outputs generally leads to higher perplexity and lower
average accuracy.
In particular, output-side memory injection produces
the highest LAMBADA perplexity of 71.83.
These observations indicate that memory is more effective
when enriching the information aggregated by attention,
rather than directly modifying attention matching or
the final attention outputs.
\begin{table*}[h]
\centering

\begin{minipage}[t]{0.26\textwidth}
\vspace{0pt}
\centering
\includegraphics[width=\linewidth]{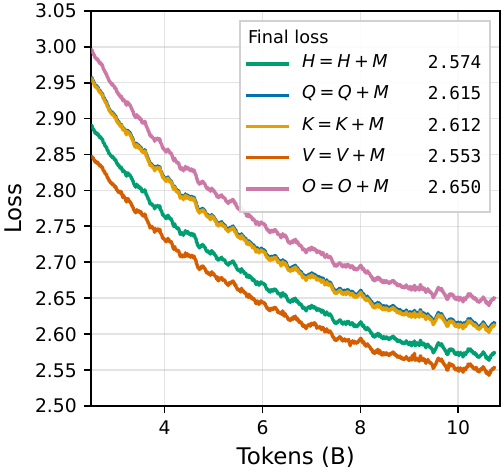}
\end{minipage}%
\hfill
\begin{minipage}[t]{0.7\textwidth}
\vspace{0pt}
\centering
\scriptsize
\renewcommand{\arraystretch}{1.2}
\setlength{\tabcolsep}{4pt}

\resizebox{\linewidth}{!}{%
\begin{tabular}{lrcc|cccccccc}
\toprule
\multirow{2}{*}{\textbf{Configuration}}
& \multirow{2}{*}{\textbf{Params.}}
& \textbf{Lamb.}
& \textbf{WikiText}
& \textbf{Lamb.}
& \textbf{ARC-E}
& \textbf{ARC-C}
& \textbf{HellaSwag}
& \textbf{PIQA}
& \textbf{WinoGrande}
& \textbf{OBQA}
& \textbf{Avg.} \\
& &
$\mathrm{PPL}\downarrow$
& $\mathrm{PPL}\downarrow$
& $\mathrm{Acc.}\uparrow$
& $\mathrm{Acc.}\uparrow$
& $\mathrm{Acc.}_{n}\uparrow$
& $\mathrm{Acc.}_{n}\uparrow$
& $\mathrm{Acc.}\uparrow$
& $\mathrm{Acc.}\uparrow$
& $\mathrm{Acc.}\uparrow$
& $\uparrow$ \\
\midrule
\multicolumn{12}{l}{%
\textit{24 layers, hidden size 1,024; 10B training tokens;
0.5M tokens per batch}
} \\
\midrule
$H = H+M$& 526M 
& 50.22 & 30.68 & 30.66 & 55.22 & 26.45 & 36.98 & 66.38 & 51.07 & 21.20 & 41.14 \\
$Q = Q+M$& 526M 
& 59.69 & 33.17 & 29.77 & 53.62 & 27.73 & 35.43 & 65.07 & 52.80 & 19.40 & 40.55 \\
$K=K+M$
& 526M & 58.23 & 32.63 & 29.21 & 53.75 & 26.19
& 35.69 & 65.56 & 49.88 & \textbf{22.20} & 40.35 \\
$V =V+M$
& 526M & 45.93 & \textbf{29.82} & \textbf{31.57} & 55.05 & 26.62
& 37.52 & 65.61 & 52.49 & 20.80 & \textbf{41.38} \\
$O=O+M$& 526M 
& 71.83 & 34.47 & 28.97 & 53.32 & 25.34 & 34.45 & 64.58 & 51.54 & 20.60 & 39.83 \\

\bottomrule
\end{tabular}%
}

\captionsetup{
    font=scriptsize,
    labelfont=normalfont,
    justification=raggedright,
    singlelinecheck=false,
    skip=4pt
}

\caption{%
Ablation of memory placement with shared keys and values ($K=V$).
Memory $M$ is added to the hidden states ($H$), queries ($Q$),
keys ($K$), values ($V$), or attention outputs ($O$).
Bold marks the best result in each metric column.
The left panel shows the corresponding training loss curves.
}
\label{tab:placement}
\end{minipage}
\vspace{-20pt}
\end{table*}

\paragraph{Memory Normalization and RoPE Placement.}
We investigate the effects of memory normalization and
the choice of key representations for value sharing.
As shown in Table~\ref{tab:normalization}, normalizing
the memory embeddings substantially improves performance,
reducing LAMBADA perplexity from 54.46 to 45.93 and
WikiText perplexity from 32.60 to 29.82, while increasing
the average downstream accuracy from 40.85 to 41.38.
This suggests that controlling the scale of memory
embeddings is important for effective memory integration.

We further examine whether values should reuse keys
before or after rotary positional embeddings (RoPE).
Using RoPE-transformed keys
($V=K^{\mathrm{RoPE}}+\mathrm{Norm}(E[i])$)
yields higher perplexity and lower average accuracy
than using pre-RoPE keys
($V=K+\mathrm{Norm}(E[i])$).
Specifically, the latter reduces LAMBADA perplexity
from 51.38 to 45.93 and improves average accuracy
from 40.61 to 41.38.
These results suggest that positional transformations
beneficial for attention matching may be unnecessary
for value representations, motivating our choice
to share keys before applying RoPE.

\begin{table*}[h]
\centering

\begin{minipage}[t]{0.26\textwidth}
\vspace{0pt}
\centering
\includegraphics[width=\linewidth]{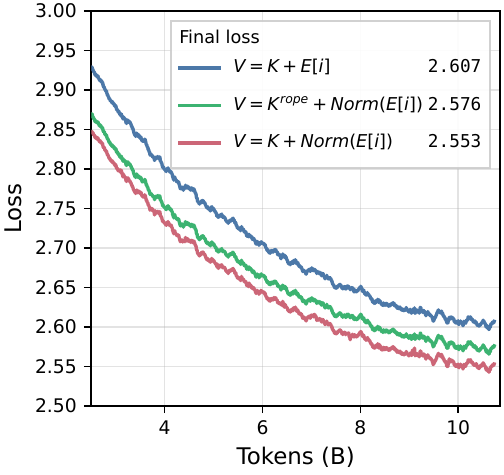}
\end{minipage}%
\hfill
\begin{minipage}[t]{0.7\textwidth}
\vspace{0pt}
\centering
\scriptsize
\renewcommand{\arraystretch}{1.2}
\setlength{\tabcolsep}{4pt}

\resizebox{\linewidth}{!}{%
\begin{tabular}{lrcc|cccccccc}
\toprule
\multirow{2}{*}{\textbf{Configuration}}
& \multirow{2}{*}{\textbf{Params.}}
& \textbf{Lamb.}
& \textbf{WikiText}
& \textbf{Lamb.}
& \textbf{ARC-E}
& \textbf{ARC-C}
& \textbf{HellaSwag}
& \textbf{PIQA}
& \textbf{WinoGrande}
& \textbf{OBQA}
& \textbf{Avg.} \\
& &
$\mathrm{PPL}\downarrow$
& $\mathrm{PPL}\downarrow$
& $\mathrm{Acc.}\uparrow$
& $\mathrm{Acc.}\uparrow$
& $\mathrm{Acc.}_{n}\uparrow$
& $\mathrm{Acc.}_{n}\uparrow$
& $\mathrm{Acc.}\uparrow$
& $\mathrm{Acc.}\uparrow$
& $\mathrm{Acc.}\uparrow$
& $\uparrow$ \\
\midrule

\multicolumn{12}{l}{%
\textit{24 layers, hidden size 1,024; 10B training tokens;
0.5M tokens per batch}
} \\
\midrule

$V=K+E[i]$
& 526M
& 54.46& 32.60& 30.22& 54.25& \textbf{26.96}& 36.01& 65.56& 52.33& 20.60& 40.85 \\

$V=K^{\mathrm{RoPE}}+\mathrm{Norm}(E[i])$
& 526M
& 51.38& 30.79& 30.18& 54.17& 26.11& 36.79& 65.23& 51.78& 20.00& 40.61 \\

$V=K+\mathrm{Norm}(E[i])$
& 526M
& \textbf{45.93}& \textbf{29.82}& \textbf{31.57}& \textbf{55.05}& 26.62& \textbf{37.52}& \textbf{65.61}& \textbf{52.49}& \textbf{20.80}& \textbf{41.38} \\

\bottomrule
\end{tabular}%
}

\captionsetup{
    font=scriptsize,
    labelfont=normalfont,
    justification=raggedright,
    singlelinecheck=false,
    skip=8pt
}

\caption{%
Ablation of memory normalization and RoPE placement
in key--value sharing.
We compare unnormalized and normalized memory embeddings,
as well as sharing key representations before ($K$)
or after ($K^{\mathrm{RoPE}}$) applying rotary positional
embeddings.
Bold indicates the best result in each metric column.
The left panel shows the corresponding training loss curves.
}
\label{tab:normalization}

\end{minipage}
\vspace{-20pt}
\end{table*}

\paragraph{Compatibility with Attention Residuals.}
Finally, we investigate whether MA is complementary to
attention residuals (AttnRes).
As shown in Table~\ref{tab:attnres},
introducing AttnRes substantially improves language
modeling perplexity for both standard attention and MA.
In particular, MA with AttnRes achieves the lowest
LAMBADA and WikiText perplexities of 38.37 and 28.88,
respectively, compared with 45.93 and 29.82 for MA alone.
It also achieves the highest average downstream accuracy
of 41.46.
These results demonstrate that MA is compatible with
AttnRes, suggesting that explicit memory augmentation
and residual information propagation can provide
complementary benefits.

\begin{table*}[h]
\centering

\begin{minipage}[t]{0.26\textwidth}
\vspace{0pt}
\centering
\includegraphics[width=\linewidth]{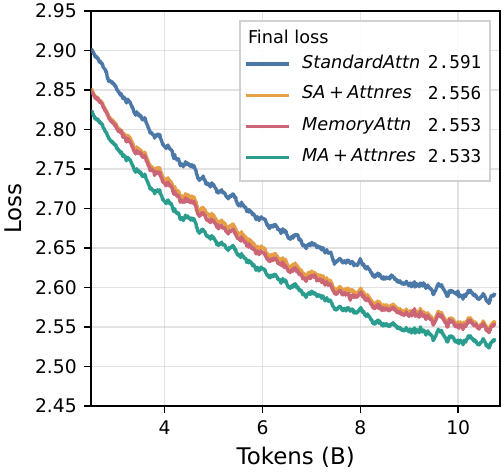}
\end{minipage}%
\hfill
\begin{minipage}[t]{0.7\textwidth}
\vspace{0pt}
\centering
\scriptsize
\renewcommand{\arraystretch}{1.2}
\setlength{\tabcolsep}{4pt}

\resizebox{\linewidth}{!}{%
\begin{tabular}{lrcc|cccccccc}
\toprule
\multirow{2}{*}{\textbf{Configuration}}
& \multirow{2}{*}{\textbf{Params.}}
& \textbf{Lamb.}
& \textbf{WikiText}
& \textbf{Lamb.}
& \textbf{ARC-E}
& \textbf{ARC-C}
& \textbf{HellaSwag}
& \textbf{PIQA}
& \textbf{WinoGrande}
& \textbf{OBQA}
& \textbf{Avg.} \\
& &
$\mathrm{PPL}\downarrow$
& $\mathrm{PPL}\downarrow$
& $\mathrm{Acc.}\uparrow$
& $\mathrm{Acc.}\uparrow$
& $\mathrm{Acc.}_{n}\uparrow$
& $\mathrm{Acc.}_{n}\uparrow$
& $\mathrm{Acc.}\uparrow$
& $\mathrm{Acc.}\uparrow$
& $\mathrm{Acc.}\uparrow$
& $\uparrow$ \\
\midrule
\multicolumn{12}{l}{%
\textit{24 layers, hidden size 1,024; 10B training tokens;
0.5M tokens per batch}
} \\
\midrule
Standard
& 336M & 50.38 & 31.84 & 31.32 & 54.17 & 26.88
& 35.98 & 64.53 & \textbf{53.28} & \textbf{21.60} & 41.11 \\
Standard+Attnres & 336M
& 39.59 & 29.77 & 32.12 & 54.71 & \textbf{27.13} & 37.47 & 65.61 & 49.33 & 20.60 & 41.00 \\
MA& 526M 
& 45.93 & 29.82 & 31.57& 55.05 & 26.62 & 37.52 & 65.61 & 52.49 & 20.80 & 41.38 \\
MA+Attnres& 526M
& \textbf{38.37} & \textbf{28.88} & \textbf{32.49} & \textbf{55.68} & 25.17 & \textbf{38.18} & \textbf{66.70} & 51.22 & 20.80 & \textbf{41.46} \\
\bottomrule
\end{tabular}%
}

\captionsetup{
    font=scriptsize,
    labelfont=normalfont,
    justification=raggedright,
    singlelinecheck=false,
    skip=4pt
}

\caption{%
Effect of AttnRes on Standard and MA models under matched training
token budgets. Bold marks the best result in each metric column.
The left panel shows the corresponding training loss curves.
}
\label{tab:attnres}
\end{minipage}

\end{table*}
\subsection{Training and Inference Efficiency}
\label{sec:efficiency}
\begin{figure}[ht]
    \centering
    \includegraphics[width=\linewidth]{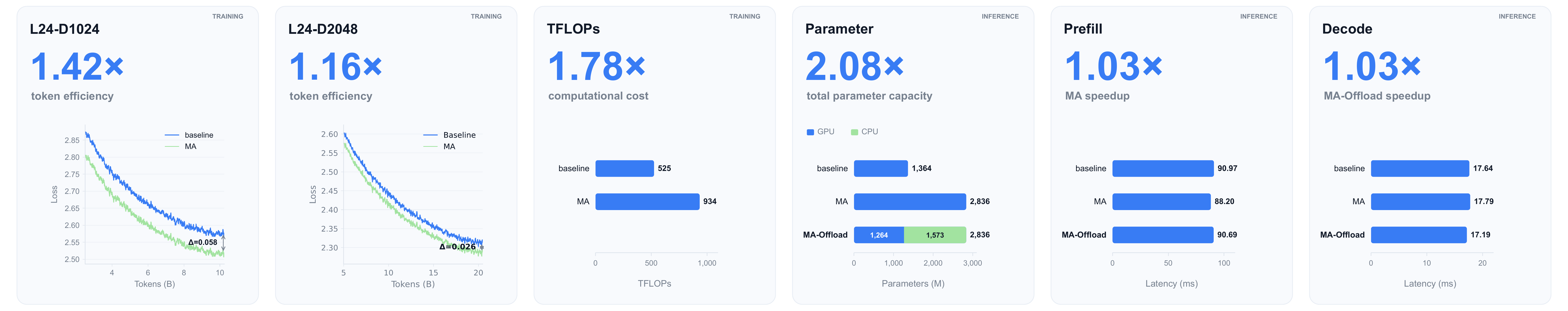}
    \caption{
        Training and inference comparison of MA and Standard. At the matched-loss operating points shown, MA achieves $1.42\times$ and $1.16\times$ token efficiency for L24-D1024 and L24-D2048, respectively. The remaining panels report training FLOPs, parameter placement, and model forward-pass latency for L24-D2048. MA-Offload stores memory-table parameters in CPU memory.
    }
    \label{fig:loss_efficiency}
\end{figure}
We examine whether MA can expand parameter capacity while reducing value-construction arithmetic and GPU parameter storage. We first compare training token efficiency, then evaluate inference latency and parameter placement. The inference results use a prototype implementation with prefetching and computation--transfer overlap.

\paragraph{Training efficiency.}
We compare the number of training tokens required to reach a matched loss. For a target loss $\ell$, let $n_{\mathrm{Standard}}(\ell)$ and $n_{\mathrm{MA}}(\ell)$ denote the corresponding token counts. We define token efficiency as
\begin{equation}
    \rho_{\mathrm{token}}(\ell)
    =
    \frac{n_{\mathrm{Standard}}(\ell)}
         {n_{\mathrm{MA}}(\ell)}.
\end{equation}
At the operating points shown in Figure~\ref{fig:loss_efficiency}, MA achieves token efficiencies of $1.42\times$ and $1.16\times$ for L24-D1024 and L24-D2048, respectively. These correspond to approximately 29.6\% and 13.8\% fewer training tokens to reach the selected loss levels. The comparisons use models with additional memory parameters and measure token efficiency rather than wall-clock training speed.

\paragraph{Inference configurations.}
We compare standard attention (\textsc{Standard}), MA with memory tables stored on the GPU (\textsc{MA}), and MA with CPU-offloaded memory tables (\textsc{MA-Offload}). Both MA variants fold head-wise RMSNorm into the tables before inference, reducing online value construction to lookup and addition. MA-Offload additionally uses a prefetch pipeline to overlap CPU retrieval and host-to-device transfer with model computation. All three configurations retain a conventional KV cache; these measurements do not include MA-Recall.

Experiments use a single NVIDIA H800 GPU with BF16 precision, PyTorch 2.9.1 (CUDA 12.6), and FlashAttention 2.8.3. The model has 24 layers, a hidden size of 2,048, 32 query heads, and 32 key/value heads. Each layer uses pre-normalized residual blocks and a gated MLP with an intermediate size of 5,632. The vocabulary contains 32,000 tokens, and the input embedding and output head are untied.

For the reference workload, we use batch size 8, prefill length 2,048, and a cached decoding history of 2,048 tokens. Each decoding measurement processes one new token per sequence. The offload group size is one layer for prefill and four layers for decoding, with a prefetch depth of four in both stages. MA performs layer-wise memory lookup.

\paragraph{Measurement scope.}
We measure the model forward pass, including the input embedding, Transformer layers, final RMSNorm, and output head. Prefill computes logits at every input position. For MA-Offload, the measurement includes memory retrieval and transfer during the forward pass. Timing excludes table initialization, normalization folding, decode-prefix construction, sampling, and token-ID transfers from GPU to CPU. Token IDs are available on both devices before measurement. The reported latencies therefore describe model forward passes rather than end-to-end serving.

Parameter-storage estimates include all parameters retained for inference, including folded memory tables, and distinguish their storage devices. They exclude KV caches, activations, and transfer buffers and therefore do not measure total or peak GPU memory usage.

\begin{table*}[t]
    \centering
    \small
    \renewcommand{\arraystretch}{1.15}
    \setlength{\tabcolsep}{6pt}
    \begin{tabular}{lrrrrrrr}
        \toprule
        \multirow{2}{*}{\textbf{Configuration}}
        & \multicolumn{2}{c}{\textbf{Latency (ms)}}
        & \multicolumn{3}{c}{\textbf{Parameters (M)}}
        & \multicolumn{2}{c}{\textbf{Parameter storage (MiB)}} \\
        \cmidrule(lr){2-3}
        \cmidrule(lr){4-6}
        \cmidrule(lr){7-8}
        & Prefill & Decode
        & GPU & CPU & Total
        & GPU & CPU \\
        \midrule
        \textsc{Standard}
        & 90.970 & 17.636
        & 1,364.298 & 0.000 & 1,364.298
        & 2,602.38 & 0.00 \\
        \textsc{MA}
        & 88.202 & 17.788
        & 2,836.498 & 0.000 & 2,836.498
        & 5,410.38 & 0.00 \\
        \textsc{MA-Offload}
        & 90.686 & 17.189
        & 1,263.634 & 1,572.864 & 2,836.498
        & 2,410.38 & 3,000.00 \\
        \bottomrule
    \end{tabular}
    \caption{
        Model forward-pass latency and parameter placement for the reference workload: batch size 8, prefill length 2,048, and one decoding step with a cached history of 2,048 tokens. Timing and parameter counts cover the complete model, including the input embedding, final normalization, and output head. Parameter storage is reported at BF16 precision and excludes KV caches, activations, and transfer buffers. M denotes $10^6$ parameters; MiB denotes $2^{20}$ bytes.
    }
    \label{tab:ma_inference}
\end{table*}

\paragraph{Value-construction overhead.}
MA replaces the dedicated value-projection matrix multiplication with memory lookup and element-wise addition. As shown in Table~\ref{tab:ma_inference}, MA reduces measured prefill latency from 90.970 to 88.202 ms (3.04\%), while decode latency increases slightly from 17.636 to 17.788 ms (0.86\%). Thus, despite increasing total parameter count to approximately $2.08\times$ that of Standard, MA maintains comparable forward-pass latency at this operating point. The results also illustrate that reduced value-construction arithmetic does not uniformly translate into lower latency, as memory access and execution overhead remain relevant.

\paragraph{GPU parameter storage.}
MA-Offload moves 1,572.864M memory-table parameters to CPU memory, corresponding to 3,000 MiB at BF16 precision. This reduces GPU parameter storage from 5,410.38 MiB for MA to 2,410.38 MiB, a reduction of 55.45\%. Relative to Standard, GPU parameter storage decreases by 192 MiB (7.38\%), reflecting the removal of the dedicated value projections. MA-Offload therefore accommodates approximately $2.08\times$ the total parameters of Standard with less GPU parameter storage, while using additional CPU memory.

\paragraph{Transfer overhead and overlap.}
Offloading introduces CPU memory access and host-to-device transfers. Since lookup addresses depend only on token IDs and layer indices, our prototype prefetches memory entries and overlaps retrieval and transfer with model computation to limit the overhead exposed on the critical path.

At the reference workload, MA-Offload records prefill and decode latencies of 90.686 and 17.189 ms, respectively, compared with 90.970 and 17.636 ms for Standard. These correspond to measured latency reductions of 0.31\% and 2.53\%. The results demonstrate that, in this configuration, the additional memory traffic can be accommodated without a net increase in measured forward-pass latency. These aggregate timings combine the effects of reduced value-projection computation, memory access, transfer, and scheduling; they do not establish that communication is cost-free or isolate the contribution of overlap.

\paragraph{Implementation scope.}
Our implementation is a prototype for examining the storage--latency trade-off. The measurements demonstrate the feasibility of expanding parameter capacity and reducing GPU parameter storage while maintaining near-baseline inference latency. Further optimization of lookup, buffering, and transfer scheduling may improve performance, but is not evaluated here. The small timing differences should be interpreted as measurements at the reported operating point rather than consistent speed advantages across workloads.

\begin{figure}[t]
    \centering
    \includegraphics[width=\linewidth]{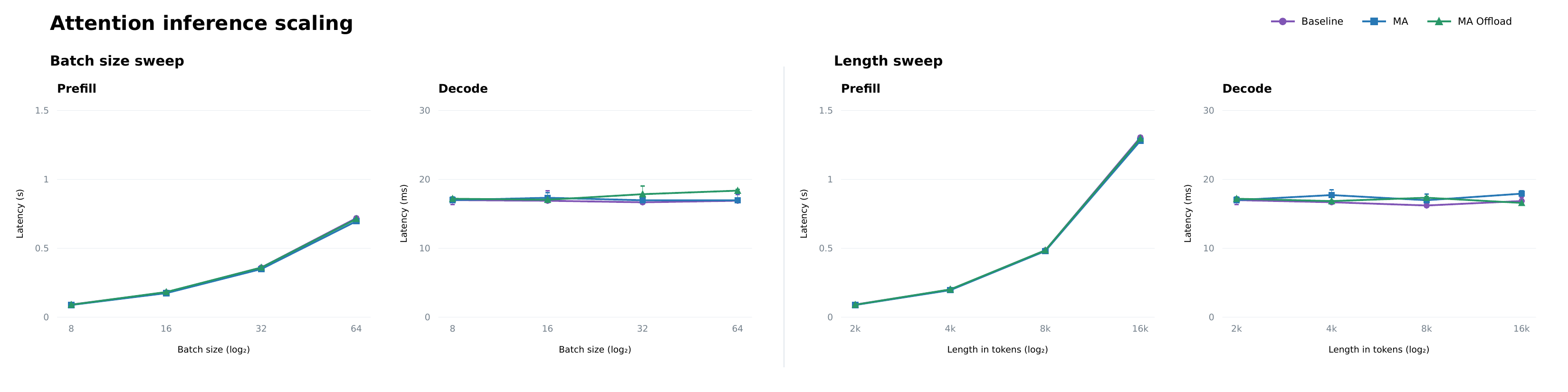}
    \caption{
        Model forward-pass latency of Standard, MA, and MA-Offload. Batch size varies at length 2K (left), and length varies at batch size 8 (right). Each pair of panels reports prefill and decoding latency. Points indicate medians; error bars show the minimum and maximum across measurement rounds. Horizontal axes use a logarithmic scale with base 2; 1K denotes 1,024 tokens. Lower is better.
    }
    \label{fig:inference_sweep}
\end{figure}

\section*{Acknowledgments}

This work was conducted during my time at Yuanshi Intelligence.
I would like to thank Huanqi Cao, Zhiyuan Li, Zhan Zheng,
Ziyin Yue, and Zhouran Ji for their valuable help and support.
I am also grateful to Jiacheng You, Jianlin Su, and other
members of the FLA community for their insightful discussions,
guidance, and constructive suggestions.

Much remains to be explored, both theoretically and
in terms of infrastructure. Memory Attention is only
a starting point, and I hope it will eventually become
a building block of future model architectures.

\newpage
\bibliography{bib}
\bibliographystyle{plain}
\appendix
\section{Pretraining Settings}
\label{sec:pretraining_settings}

\paragraph{Data and optimization.}
We conduct pretraining experiments on FineWeb-10BT. Unless otherwise specified, models are trained for 20,480 optimization steps with a sequence length of 2,048 and a global batch size of 256 sequences. We use AdamW with a peak learning rate of $3\times10^{-4}$ and $\epsilon=10^{-15}$. The learning rate is warmed up over the first 1,024 steps and then decayed using a cosine schedule to 10\% of its peak value. The global gradient norm is clipped at 1.0. Within each backbone configuration, Standard and MA use the same training settings and random seed for controlled comparison. Table~\ref{tab:pretrain_settings} summarizes the default training hyperparameters.

\begin{table}[ht]
    \centering
    \small
    \renewcommand{\arraystretch}{1.15}
    \setlength{\tabcolsep}{8pt}
    \begin{tabular}{lc}
        \toprule
        \textbf{Hyperparameter} & \textbf{Value} \\
        \midrule
        Dataset & FineWeb-10BT \\
        Optimizer & AdamW \\
        Peak learning rate & $3\times10^{-4}$ \\
        Adam $\epsilon$ & $10^{-15}$ \\
        Learning-rate schedule & Cosine decay \\
        Warmup steps & 1,024 \\
        Minimum learning-rate ratio & 0.1 \\
        Global batch size (sequences) & 256 \\
        Sequence length & 2,048 \\
        Training steps & 20,480 \\
        Gradient clipping threshold & 1.0 \\
        Random seed & 42 \\
        \bottomrule
    \end{tabular}
    \caption{
        Default pretraining hyperparameters. The minimum learning-rate ratio is relative to the peak learning rate.
    }
    \label{tab:pretrain_settings}
\end{table}

\paragraph{Backbone and attention configurations.}
For the hidden-dimension-1,024 backbone, all models use 24 Transformer layers. MHA uses 16 query heads and 16 key/value heads, while GQA uses 16 query heads and eight key/value heads; both use a head dimension of 64. MQA uses four query heads sharing a single key/value head, with a head dimension of 256. Each key/value head is therefore associated with one, two, and four query heads in MHA, GQA, and MQA, respectively. Keys and values have matching per-head dimensions within each configuration. The models use RoPE with a base of 10,000, RMSNorm with $\epsilon=10^{-6}$, and untied input embeddings and output heads. Table~\ref{tab:attention_head_settings} summarizes these configurations. Standard and MA use identical head settings within each comparison; comparisons across attention configurations also reflect differences in head count and dimension.

\begin{table}[ht]
    \centering
    \small
    \renewcommand{\arraystretch}{1.15}
    \setlength{\tabcolsep}{6pt}
    \begin{tabular}{lccc}
        \toprule
        \textbf{Configuration} & \textbf{MHA} & \textbf{GQA} & \textbf{MQA} \\
        \midrule
        Number of layers & 24 & 24 & 24 \\
        Hidden dimension & 1,024 & 1,024 & 1,024 \\
        Query heads ($H_Q$) & 16 & 16 & 4 \\
        KV heads ($H_{\mathrm{KV}}$) & 16 & 8 & 1 \\
        KV head dimension ($d_h$) & 64 & 64 & 256 \\
        Query heads per KV head & 1 & 2 & 4 \\
        Total KV dimension ($d_v$) & 1,024 & 512 & 256 \\
        RoPE base & 10,000 & 10,000 & 10,000 \\
        RMSNorm $\epsilon$ & $10^{-6}$ & $10^{-6}$ & $10^{-6}$ \\
        Tied input/output embeddings & No & No & No \\
        \bottomrule
    \end{tabular}
    \caption{
        Attention configurations for the 24-layer backbone with hidden dimension 1,024. KV head dimension denotes the dimension of each key/value head, while $d_v=H_{\mathrm{KV}}d_h$ denotes the total dimension of keys or values across heads. Standard and MA use the same head configuration within each comparison.
    }
    \label{tab:attention_head_settings}
\end{table}

\paragraph{Memory configuration.}
Each MA layer maintains a token memory table $\mathbf{E}\in\mathbb{R}^{N\times d_v}$, where $N$ is the vocabulary size and $d_v=H_{\mathrm{KV}}d_h$. The memory-table widths are therefore 1,024, 512, and 256 for MHA, GQA, and MQA, respectively. Retrieved memory vectors are normalized independently within each key/value head and added to the corresponding keys before RoPE to construct values. Query heads sharing a key/value head also share its memory contribution. Memory tables are independent across layers and are learned jointly with the rest of the model.


\end{document}